# FLEURS-R: A Restored Multilingual Speech Corpus for Generation Tasks

*Min Ma[1], Yuma Koizumi[2], Shigeki Karita[2], Heiga Zen[2], Jason Riesa[1], Haruko Ishikawa[2], Michiel Bacchiani[2]*

[1] Google DeepMind, USA    [2] Google DeepMind, Japan

{minm, koizumiyuma, heigazen}google.com

## Abstract

This paper introduces FLEURS-R, a speech restoration applied version of the Few-shot Learning Evaluation of Universal Representations of Speech (FLEURS) corpus. FLEURS-R maintains an N-way parallel speech corpus in 102 languages as FLEURS, with improved audio quality and fidelity by applying the speech restoration model Miipher. The aim of FLEURS-R is to advance speech technology in more languages and catalyze research including text-to-speech (TTS) and other speech generation tasks in low-resource languages. Comprehensive evaluations with the restored speech and TTS baseline models trained from the new corpus show that the new corpus obtained significantly improved speech quality while maintaining the semantic contents of the speech. The corpus is publicly released via Hugging Face[1].

## 1. Introduction

There have been rapid development in the speech generation area in the past few years. Models such as denoising diffusion probabilistic models (DDPMs) [1, 2], neural audio codec [3, 4], and large language models (LLMs) [5–7] have been successfully applied to speech generation tasks. As these speech generation tasks can be viewed as a sequence-to-sequence generative task, the progress in generative models in different areas can be introduced to improve the performance. Now naturally sounding synthetic speech with an arbitrary speaker's voice can be synthesized with a small amount of speech data [8]. There are models which support controlling their speaking styles and/or voice characteristics via natural language-based prompts [9, 10].

Although there have been great advancements in the modeling side, the progress in the data side is relatively slow. As these new generative models are language agnostic and can be pre-trained from a large quantity of speech-only or text-only data (*e.g.*, 100k hours) [5, 11], the required amount of speech-text paired data is getting smaller [7]. This nature is highly relevant for developing multilingual speech generation models, especially for low-resource languages.

The FLEURS [12] corpus covers 102 languages, which spans over 17 language families and 27 unique writing systems. It was designed to enable speech technology in more languages and catalyze research in low-resource speech understanding. However, as all recordings are kept as they-are, either from quiet or noisy environment, making it less ideal for speech generation tasks, where models are requested to produce high-quality speech.

Recently, Koizumi *et al.* introduced LibriTTS-R [13], which is a speech restoration applied version of the LibriTTS corpus [14]. As it offers speech signals in higher sampling rate, less noise, and less reverberation, neural end-to-end TTS models trained with LibriTTS-R achieved better subjective naturalness than those with the original LibriTTS corpus [13].

This paper introduces FLEURS-R, a speech-restoration applied version of the FLEURS corpus. It keeps the same properties as the original FLEURS corpus with improved audio quality, *i.e.*, less noise and reverberation with higher sampling rate (24 kHz). Table 1 compares FLEURS-R with existing common public multilingual TTS corpora. The key properties of FLEURS-R that are:

- Containing N-way parallel speech and text in 102 languages; the improved speech quality makes it a better choice for speech generation tasks, including TTS, speech-to-speech translation (S2ST) and voice conversion (VC).

- Highly multilingual (102 languages) where 80% languages are low-resource. It helps catalyze speech generation research in multilingual, cross-lingual and low-resource settings.

## 2. Speech Restoration Pipeline

### 2.1. Speech Restoration Model

We restored the FLEURS speech samples using the same methodology employed to create the LibriTTS-R corpus [13]. LibriTTS-R was created by applying a speech restoration model *Miipher* [19] to the LibriTTS corpus [14]. Miipher extracts acoustic features from noisy speech using w2v-BERT [20]. The system then employs DF-Conformer [21] to convert these noisy acoustic features into clean ones while using speaker and text conditioning features extracted by speaker-encoder [22] and PnG-BERT [23]. Finally, the WaveFit [24] neural vocoder generates a clean speech waveform from the predicted features.

Since FLEURS is a multilingual corpus and Miipher supports only English [19], we made several updates to the Miipher model structure to accommodate this difference. First, we replaced the acoustic feature extractor from w2v-BERT [20] to the Universal Speech Model (USM) [25]. Unlike w2v-BERT [20], which was trained on English speech samples, the USM was pre-trained on a massive dataset of 12 million hours of speech spanning over 300 languages. We used a non-fine-tuned USM encoder to preserve the speaker's acoustic characteristics in the extracted features. Specifically, we used the 2-billion parameter "pre-trained" model [25]. In self-supervised learning (SSL) speech feature extraction, it is known that deeper layers tend to lose detailed and local acoustic information [26]; therefore, we used the intermediate feature from the 13th of 32th layers based on preliminary experiments.

---

[1] https://huggingface.co/datasets/google/fleurs-r

Table 1: *Comparison among FLEURS-R and other common public speech corpora.*

| Dataset | #Locales | Total Duration | Domains | Speech Type | Sampling Rate | License | Parallel speech |
|---|---|---|---|---|---|---|---|
| MLS [15] | 8 | 50.5k hours | Audiobook | Read | 16 kHz | CC-BY-4.0 [16] | No |
| CML-TTS [17] | 7 | 3.2k hours | Audiobook | Read | 24 kHz | CC-BY-4.0 [16] | No |
| M-AILabs speech datasets[2] | 9 | 1k hours | Audiobook | Read | 16 kHz | BSD 3-Clause License | No |
| BC2013 [18] | 4 | 0.3k hours | Audiobook | Read | 44.1 kHz | Non-commercial | No |
| LibriTTS-R [13] | 1 | 0.6k hours | Audiobook | Read | 24 kHz | CC-BY-4.0 [16] | No |
| FLEURS [12] | 102 | 1.4k hours | Wikipedia | Read | 16 kHz | CC-BY-4.0 [16] | Yes |
| FLEURS-R (this work) | 102 | 1.3k hours | Wikipedia | Read | 24 kHz | CC-BY-4.0 [16] | Yes |

Furthermore, preliminary experiments indicated that the USM features retained both acoustic details; neither text nor speaker conditioning improved the reconstruction accuracy. Consequently, both speaker encoder [22] and PnG-BERT text encoder [23] were removed from the new Miipher network architecture.

### 2.2. Data Processing Pipeline

First we applied the new Miipher model-based speech restoration to the complete set in the original FLEURS corpus. Thanks to Miipher's audio super-resolution capability, the sampling rate of speech samples in the FLEURS-R samples was increased from 16 kHz to 24 kHz. Note that FLEURS-R maintains the same constituent samples as the original FLEURS corpus except the audio quality.

Due to possible errors caused in the Miipher speech restoration process, some restored samples may exhibit signal processing artifacts. To identify successfully restored samples, we performed ASR-based filtering. The list of the rejected samples will also be published. Note that all experiments in Sec. 3 including TTS model training were conducted with the rejected samples.

## 3. Evaluations

We conducted ASR-based intelligibility evaluations, automatic subjective naturalness evaluations, and TTS model training experiments with the new FLEURS-R corpus. Some demo samples from each experiment are available as a supplementary material to honor double-blind review.

### 3.1. ASR Evaluation

To validate the consistency of semantic contents between original FLEURS and new FLEURS-R corpora, we conducted ASR evaluations over all 102 languages. We used the Maestro-U [27] grapheme ASR model which performed reasonably well in terms of character error rates (CERs) in most of these 102 languages.

Table 2 shows the language-specific CERs. The abbreviations in the first row denote regions; Western European (WE), Eastern European (EE), Central-Asia, Middle-East and North-Africa (CMN), Sub-Saharan Africa (SSA), South-Asia (SA), South-East Asia (SEA), and Chinese, Japanese and Korean (CJK). Please refer to [12] for the individual locale codes. It can be seen from the table that the average CERs over all locales for FLEURS and FLEURS-R were approximately equal (9.67% and 9.74%). This suggests the speech restoration process maintained the semantic contents in the original speech in most languages. 32% languages got improved CERs, especially in Xhosa (xh), Umbundu (umb), Macedonian (mk), Tamil (ta), Turkish (tr), Hebrew (he) and Armenian (hy). The gains mostly come from the reduced substitution error rates, likely because enhanced speech quality makes the acoustically similar words more distinguishable. Other locales maintained or only saw small regressions in CERs. The exceptional locales were Nepali (ne), Punjabi (pa), Indonesian (id), Latvian (lv), and Czech (cs). Such degradation were from higher substitution and deletion error rates. On most locales, insertion error rates were generally reduced, indicating that Miipher reduced the environment noises of speech recordings. Three locales, Sorani-Kurdish (ckb), Sindhi (sd) and Cantonese (yue) observed high CERs, though their ASR baseline CERs on FLEURS are already high. We provide samples from the two groups (most improved and most regressed) in the supplementary materials.

Table 2: *The character error rates (%) for both FLEURS (top) and FLEURS-R (bottom) corpora in all 102 languages.*

| WE | | | | | | | | | | | | | |
|---|---|---|---|---|---|---|---|---|---|---|---|---|---|
| ast | bs | ca | hr | da | nl | en | fi | fr | gl | de | el | hu | is | ga |
| 4.6 | 3.0 | 2.9 | 3.5 | 6.7 | 3.3 | 5.7 | 2.0 | 4.8 | 2.6 | 2.5 | 5.0 | 6.9 | 6.6 | 23.8 |
| 4.9 | 3.1 | 3.1 | 3.5 | 8.0 | 3.2 | 5.7 | 2.3 | 4.7 | 2.6 | 2.6 | 5.7 | 6.3 | 6.8 | 24.4 |

| WE | | | | | | | | | EE | | | | |
|---|---|---|---|---|---|---|---|---|---|---|---|---|---|
| it | kea | lb | mt | nb | oc | pt | es | sv | cy | am | be | bg | cs | et |
| 1.5 | 4.5 | 6.3 | 3.2 | 4.4 | 8.3 | 3.0 | 1.9 | 4.1 | 7.5 | 9.1 | 3.4 | 2.7 | 3.5 | 2.1 |
| 1.6 | 4.6 | 6.6 | 3.3 | 4.6 | 9.3 | 3.0 | 1.9 | 4.5 | 7.6 | 9.4 | 3.4 | 2.9 | 4.2 | 2.0 |

| EE | | | | | | | | | CMN | | | | |
|---|---|---|---|---|---|---|---|---|---|---|---|---|---|
| ka | lv | lt | mk | pl | ro | ru | sr | sk | sl | uk | ar | az | he | kk |
| 5.7 | 2.2 | 4.1 | 2.2 | 2.6 | 3.7 | 3.1 | 11.1 | 2.1 | 4.4 | 3.3 | 6.6 | 5.8 | 17.6 | 3.6 |
| 5.6 | 2.7 | 4.4 | 1.8 | 2.7 | 3.4 | 3.2 | 11.2 | 2.0 | 4.4 | 3.1 | 6.8 | 5.4 | 15.5 | 3.2 |

| CMN | | | | | | | SSA | | | | | | |
|---|---|---|---|---|---|---|---|---|---|---|---|---|---|
| ky | mn | ps | fa | ckb | tg | tr | uz | af | am | ff | lg | ha | ig | kam |
| 4.7 | 8.9 | 17.3 | 5.2 | 74.5 | 4.5 | 4.4 | 7.6 | 5.8 | 9.1 | 11.0 | 8.9 | 7.9 | 13.2 | 12.5 |
| 4.3 | 9.7 | 17.5 | 4.9 | 78.1 | 4.4 | 3.9 | 7.9 | 5.6 | 9.4 | 11.4 | 8.9 | 7.5 | 12.4 | 11.9 |

| SSA | | | | | | | | SA | | | | | |
|---|---|---|---|---|---|---|---|---|---|---|---|---|---|
| ln | luo | nso | ny | om | sn | so | sw | umb | wo | xh | yo | zu | as | bn |
| 5.0 | 5.0 | 7.4 | 5.9 | 15.3 | 3.7 | 14.3 | 3.8 | 17.9 | 15.3 | 14.3 | 22.6 | 5.8 | 8.8 | 6.3 |
| 5.0 | 5.2 | 7.7 | 6.2 | 15.3 | 4.3 | 14.5 | 3.9 | 10.9 | 14.9 | 8.1 | 22.6 | 6.3 | 9.1 | 6.5 |

| SA | | | | | | | | | | | SEA | | |
|---|---|---|---|---|---|---|---|---|---|---|---|---|---|
| gu | hi | kn | ml | mr | ne | or | pa | sd | ta | te | ur | my | ceb | tg |
| 5.6 | 6.2 | 5.1 | 4.8 | 7.4 | 9.7 | 7.6 | 6.8 | 72.1 | 12.2 | 7.3 | 8.2 | 13.8 | 4.7 | 4.5 |
| 5.9 | 6.7 | 4.9 | 4.6 | 7.7 | 14.3 | 8.3 | 10.1 | 74.4 | 10.1 | 7.9 | 8.6 | 15.0 | 4.7 | 4.4 |

| SEA | | | | | | CJK | | | | ALL |
|---|---|---|---|---|---|---|---|---|---|---|
| id | jv | km | lo | ms | mi | th | vi | yue | cmn | ja | ko | |
| 3.3 | 5.0 | 17.8 | 22.2 | 4.0 | 10.6 | 11.1 | 14.4 | 34.8 | 27.2 | 25.0 | 15.6 | **9.2** |
| 4.0 | 5.3 | 17.0 | 23.1 | 4.6 | 12.4 | 10.9 | 14.2 | 32.1 | 27.8 | 24.6 | 14.9 | **9.2** |

### 3.2. Speech Naturalness Evaluation

While the subjective 5-point Mean Opinion Score (MOS) is a standard evaluation method to assess the naturalness, it poses

---

[2]There are multiple releases of 9 locales by M-AILabs, namely, German, Queen's English, American English, Spanish, Italian, Ukrainian, Russian, French, and Polish. We count them together for number of languages and total duration. https://www.caito.de/2019/01/03/the-m-ailabs-speech-dataset

Table 3: *The 5-scale SQuId MOS in naturalness for both FLEURS (top) and FLEURS-R (bottom) corpora in all 102 languages.*

| WE | | | | | | | | | | | | | | |
|---|---|---|---|---|---|---|---|---|---|---|---|---|---|---|
| ast | bs | ca | hr | da | nl | en | fi | fr | gl | de | el | hu | is | ga |
| 3.77 | 3.90 | 3.75 | 3.87 | 3.67 | 3.80 | 3.51 | 3.79 | 3.80 | 3.92 | 3.83 | 3.74 | 3.71 | 3.52 | 3.79 |
| 3.94 | 3.95 | 3.96 | 4.10 | 3.79 | 3.91 | 3.82 | 3.85 | 3.91 | 4.07 | 3.93 | 3.93 | 3.95 | 3.76 | 3.75 |

| WE | | | | | | | | | EE | | | | | |
|---|---|---|---|---|---|---|---|---|---|---|---|---|---|---|
| it | kea | lb | mt | nb | oc | pt | es | sv | cy | am | be | bg | cs | et |
| 3.80 | 3.66 | 3.87 | 3.86 | 3.92 | 3.80 | 3.70 | 3.93 | 3.81 | 3.71 | 4.09 | 3.77 | 3.93 | 3.69 | 3.99 |
| 3.90 | 3.92 | 3.95 | 3.98 | 3.95 | 3.60 | 4.09 | 4.07 | 3.92 | 3.79 | 4.10 | 3.86 | 3.99 | 3.83 | 4.00 |

| EE | | | | | | | | | CMN | | | | | |
|---|---|---|---|---|---|---|---|---|---|---|---|---|---|---|
| ka | lv | lt | mk | pl | ro | ru | sr | sk | sl | uk | ar | az | he | kk |
| 3.93 | 3.85 | 3.74 | 3.91 | 3.76 | 3.63 | 3.69 | 3.86 | 3.89 | 3.97 | 3.84 | 4.03 | 3.73 | 3.84 | 3.75 |
| 4.05 | 3.96 | 3.86 | 4.03 | 3.83 | 4.04 | 3.87 | 4.01 | 3.96 | 4.06 | 3.95 | 4.15 | 4.00 | 4.03 | 3.86 |

| CMN | | | | | | | SSA | | | | | | | |
|---|---|---|---|---|---|---|---|---|---|---|---|---|---|---|
| ky | mn | ps | fa | ckb | tg | tr | uz | af | am | ff | lg | ha | ig | kam |
| 3.89 | 3.77 | 3.85 | 3.72 | 3.80 | 3.95 | 3.85 | 3.71 | 3.69 | 4.09 | 3.48 | 3.82 | 3.58 | 3.27 | 3.44 |
| 3.94 | 3.93 | 3.89 | 3.94 | 3.97 | 4.04 | 4.07 | 3.94 | 3.87 | 4.10 | 3.70 | 3.90 | 3.69 | 3.56 | 3.82 |

| SSA | | | | | | | | | | | | SA | | |
|---|---|---|---|---|---|---|---|---|---|---|---|---|---|---|
| ln | luo | nso | ny | om | sn | so | sw | umb | wo | xh | yo | zu | as | bn |
| 3.29 | 3.62 | 3.27 | 3.45 | 3.83 | 3.38 | 3.50 | 3.57 | 3.25 | 3.15 | 3.48 | 3.37 | 3.54 | 3.77 | 3.64 |
| 3.55 | 3.93 | 3.49 | 3.74 | 3.99 | 3.68 | 3.77 | 3.82 | 3.62 | 3.47 | 3.72 | 3.52 | 3.58 | 3.99 | 4.07 |

| SA | | | | | | | | | | | | SEA | | |
|---|---|---|---|---|---|---|---|---|---|---|---|---|---|---|
| gu | hi | kn | ml | mr | ne | or | pa | sd | ta | te | ur | my | ceb | tg |
| 3.86 | 3.79 | 3.78 | 3.84 | 3.70 | 3.41 | 3.61 | 3.74 | 3.66 | 3.76 | 3.80 | 4.01 | 3.62 | 3.79 | 3.95 |
| 4.22 | 4.08 | 4.16 | 4.13 | 3.94 | 3.83 | 4.15 | 4.16 | 4.06 | 4.07 | 4.02 | 4.18 | 3.71 | 4.10 | 4.04 |

| SEA | | | | | | | CJK | | | | ALL |
|---|---|---|---|---|---|---|---|---|---|---|---|
| id | jv | km | lo | ms | mi | th | vi | yue | cmn | ja | ko | |
| 3.59 | 3.67 | 3.63 | 3.71 | 3.57 | 3.14 | 3.97 | 3.52 | 3.87 | 3.92 | 3.61 | 3.85 | **3.72** |
| 3.92 | 3.80 | 3.88 | 3.90 | 3.96 | 3.40 | 4.11 | 3.78 | 3.95 | 3.92 | 3.96 | 3.99 | **3.92** |

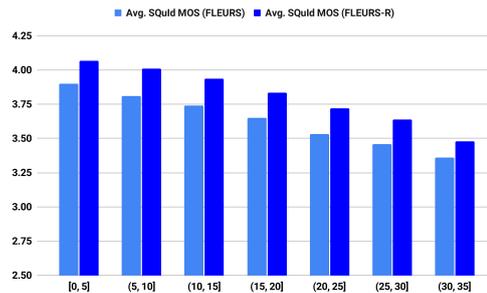

Figure 1: *Restoration brings gains in speech naturalness (in SQuId MOS), across all utterance duration ranges (in second).*

### 3.3. TTS Evaluations

#### 3.3.1. Model Configurations

We adopted the model configuration from Virtuoso 2 [29] to build a TTS baseline. It is a non-autoregressive TTS model, which uses UTF-8 byte as input representation. It aims to build a robust model supporting high and low resource languages via self-supervised and semi-supervised learning from speech-only, text-only, and speech-text pair datasets. Its speech encoder and shared encoder were composed of 6 and 18 Conformer [30] layers, respectively, with a hidden dimension of 768. The feed-forward text encoder consisted of 12 Conformer layers with a hidden dimension of 768. The semantic feature decoder in this model comprised 6 lightweight convolutional layers. The model was conditioned on both speaker and language IDs during training, allowing both speaker and language control at inference. To capture intra-speaker prosodic variations that occur in natural speech, this model has a global variational autoencoder (VAE) over input speech, which can be used to add prosodic diversity at the inference stage. Please refer to [29] for details.

We trained multi-speaker Virtuoso 2 models on either FLEURS to produce speech of 16 kHz sample rate, or trained on FLEURS-R to generate speech of 24 kHz sample rate. The same hyper-parameters were used between two models for consistent comparisons.

#### 3.3.2. Speech Naturalness Evaluation

We evaluated the naturalness of speech synthesized by the Virtuoso models by the same SQuId model. Table 4 indicates that the TTS model trained by FLEURS-R produced more natural-sounding speech, with an overall score of 3.89 compared to 3.79 for that model trained by FLEURS. In the same manner, since the speech predicted by TTS model trained on FLEURS-R has to be resampled from 24 kHz to 16 kHz before scoring, the actual naturalness of the synthesized speech should surpass than what the rating of 3.89 suggests. The most significant improvements were observed in Khmer (km), Burmese (my), Mandarin (cmn), and several South Asian languages, including Oriya (or), Hindi (hi), and Tamil (ta). We hypothesize the gains might due to shared acoustic-prosodic properties among Southeastern Asian languages, and Southern Asian languages.

#### 3.3.3. ASR Evaluation on Synthesized Speech

To evaluate the intelligibility of the synthesized speeches, we reused the same pre-existing ASR model to compute CERs on them. As shown in Table 5, the overall CERs remained consistent between two models. This suggests that the TTS models respectively trained on the restored / original corpus, could pro-challenges to evaluate the FLEURS-R corpus. As this corpus has large linguistic variations, contains 102 languages, and 80% of these languages are low-resource, it is difficult to conducting large-scale subjective evaluations. Therefore, we utilized the SQuId (Speech Quality Identifier) model [28], which was trained to predict a 5-scale subjective MOS in naturalness given an audio. It is known that SQuId doesn't map perfectly to subjective MOS and is less sensitive to linguistic correctness since the model has largely seen ratings for high-quality TTS samples (ranging between 3.0 and 5.0). However, it is still useful for relative comparisons between samples in the same language [29].

Table 3 shows that SQuId MOS in FLEURS-R were generally higher than FLEURS across all languages. On average, FLEURS-R had a 0.2 point improvement over FLEURS in the SQuId MOS. Although improvements in the SQuId MOS were observed in almost all languages except Irish (ga). Languages spoken in South Asia exhibited large gains in the SQuId MOS. Since SQuId MOS model trained on 16 kHz speech, the 24 kHz speech generated by TTS model built on FLEURS-R was resampled to 16 kHz before scoring, therefore, the actual gains in naturalness from training TTS models on FLEURS-R would be larger than the SQuId MOS score differences indicate.

We also investigated whether these score improvements were independent of utterance duration. As shown in Figure 1, the speech restored by Miipher is consistently better than original FLEURS speech in naturalness, in term of SQuId MOS. The restoration gains are especially significant on shorter utterances.

Table 4: *The 5-scale SQuId MOS in naturalness for synthetic speech generated by the Virtuoso models trained on the FLEURS (top) and the FLEURS-R (bottom) corpora in all 102 languages.*

| WE | | | | | | | | | | | | | | |
|---|---|---|---|---|---|---|---|---|---|---|---|---|---|---|
| ast | bs | ca | hr | da | nl | en | fi | fr | gl | de | el | hu | is | ga |
| 3.86 | 3.99 | 3.93 | 3.86 | 3.98 | 3.89 | 3.88 | 3.95 | 3.85 | 3.79 | 3.71 | 3.85 | 3.75 | 3.99 | 4.06 |
| 3.89 | 3.99 | 4.00 | 2.95 | 4.04 | 4.10 | 3.90 | 4.10 | 3.94 | 4.05 | 3.85 | 3.77 | 4.13 | 4.03 | 4.08 |

| WE | | | | | | | | | EE | | | | | |
|---|---|---|---|---|---|---|---|---|---|---|---|---|---|---|
| it | kea | lb | mt | nb | oc | pt | es | sv | cy | am | be | bg | cs | et |
| 3.47 | 3.87 | 3.85 | 4.00 | 3.35 | 3.61 | 3.82 | 3.91 | 3.51 | 3.70 | 3.78 | 3.08 | 3.79 | 3.83 | 3.80 |
| 3.60 | 4.07 | 3.91 | 3.95 | 3.88 | 3.46 | 3.86 | 4.06 | 3.56 | 4.13 | 3.82 | 3.89 | 3.96 | 4.01 | 3.90 |

| EE | | | | | | | | | | CMN | | | | |
|---|---|---|---|---|---|---|---|---|---|---|---|---|---|---|
| ka | lv | lt | mk | pl | ro | ru | sr | sk | sl | uk | ar | az | he | kk |
| 2.94 | 3.68 | 3.54 | 3.57 | 3.97 | 3.98 | 3.99 | 3.94 | 3.84 | 3.94 | 3.99 | 3.90 | 3.91 | 3.97 | 3.68 |
| 3.04 | 3.77 | 3.87 | 3.59 | 4.29 | 4.00 | 4.04 | 3.96 | 3.92 | 3.87 | 3.95 | 4.18 | 4.01 | 4.07 | 3.86 |

| CMN | | | | | | | | | | | SSA | | | |
|---|---|---|---|---|---|---|---|---|---|---|---|---|---|---|
| ky | mn | ps | fa | ckb | tg | tr | uz | af | am | ff | lg | ha | ig | kam |
| 3.91 | 3.73 | 3.93 | 3.97 | 3.83 | 3.86 | 3.87 | 3.91 | 3.93 | 3.78 | 3.85 | 3.86 | 3.80 | 3.47 | 4.02 |
| 4.17 | 3.98 | 3.96 | 4.07 | 4.01 | 3.76 | 3.98 | 4.00 | 3.97 | 3.82 | 3.87 | 3.80 | 4.10 | 2.96 | 4.06 |

| SSA | | | | | | | | | | | | SA | | |
|---|---|---|---|---|---|---|---|---|---|---|---|---|---|---|
| ln | luo | nso | ny | om | sn | so | sw | umb | wo | xh | yo | zu | as | bn |
| 3.66 | 3.48 | 3.80 | 3.52 | 3.64 | 3.81 | 3.83 | 3.81 | 3.08 | 3.66 | 3.87 | 3.58 | 3.60 | 3.76 | 3.90 |
| 3.69 | 3.61 | 3.94 | 3.51 | 3.46 | 3.89 | 4.03 | 3.85 | 3.15 | 3.90 | 3.61 | 3.66 | 3.60 | 4.12 | 4.01 |

| SA | | | | | | | | | | | | SEA | | |
|---|---|---|---|---|---|---|---|---|---|---|---|---|---|---|
| gu | hi | kn | ml | mr | ne | or | pa | sd | ta | te | ur | my | ceb | tg |
| 3.65 | 3.81 | 3.91 | 3.95 | 3.85 | 3.64 | 3.92 | 3.61 | 3.78 | 3.89 | 3.84 | 3.93 | 3.85 | 3.84 | 3.86 |
| 3.72 | 3.92 | 3.94 | 4.07 | 3.97 | 3.82 | 3.89 | 4.00 | 3.97 | 4.27 | 4.19 | 4.08 | 3.88 | 3.81 | 3.76 |

| SEA | | | | | | | | CJK | | | ALL |
|---|---|---|---|---|---|---|---|---|---|---|---|
| id | jv | km | lo | ms | mi | th | vi | yue | cmn | ja | ko |
| 3.93 | 3.95 | 3.90 | 3.68 | 3.98 | 3.87 | 3.70 | 3.95 | 3.70 | 3.98 | 3.84 | 3.05 | **3.79** |
| 3.94 | 4.03 | 3.87 | 3.62 | 4.22 | 3.86 | 3.80 | 4.04 | 4.11 | 3.98 | 3.83 | 3.83 | **3.89** |

duce speech with similar semantic contents. Although, there is a noticeable degradation in term of CER for the TTS generated speech (15.9%) in Table 5, *w.r.t.* original speech (9.2%) in Table 2. This difference likely results from several factors. First, the quality of the synthetic speech was still not as good as that of the natural speech. This can lead to worse ASR performance as the vast majority of training data for these ASR models consists of real speech. It is observed that the locales with extremely high CERs suffered primarily from deletion errors (e.g. Sorani-Kurdish, Sindhi, Panjabi, Japanese) or dominating substitution errors (e.g. Serbian, Madanrin, Cantonese). Second, the CER differences between them arise from both data and TTS modeling aspects. The minor CER changes (15.9% *vs.* 16.0%) between the same TTS model on different data (FLEURS, FLEURS-R) imply that the degradation is not due to data restoration. Instead, the gap indicates that the TTS models need to learn to generate speech more close to the natural speech. Figure 2 illustrates detailed error rate changes for languages where CER degraded by at least 10%. In languages like Panjabi, Serbian, Mandarin, Yoruba, and Thai, most errors resulted from substitutions. Japanese, Afrikaans, Sorani-Kurdish, and Occitan, on the other hand, experienced errors primarily due to deletions. Potential solutions include developing ASR models specifically optimized for languages with large vocabularies (like Mandarin). Additionally, adapting ASR models to better handle the acoustic conditions of the FLEURS-R dataset could help improve its performance as an estimator of intelligibility of synthetic speeches.

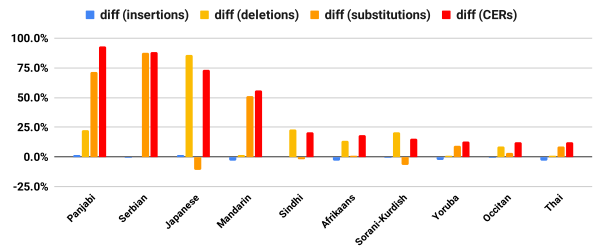

Figure 2: *Changes in three error types and CERs between ASR baselines of FLEURS, and of the predicted speech by Virtuoso model trained on FLEURS.*

Table 5: *ASR results for all the 102 languages, on synthesized speech by TTS models which are trained on original FLEURS (top) and FLEURS-R (bottom).*

| WE | | | | | | | | | | | | | | |
|---|---|---|---|---|---|---|---|---|---|---|---|---|---|---|
| ast | bs | ca | hr | da | nl | en | fi | fr | gl | de | el | hu | is | ga |
| 5.6 | 4.7 | 7.8 | 4.3 | 10.0 | 5.0 | 6.9 | 4.8 | 5.9 | 4.2 | 4.1 | 9.1 | 4.3 | 14.7 | 21.6 |
| 5.7 | 8.0 | 7.8 | 4.2 | 10.0 | 5.0 | 6.9 | 4.7 | 5.9 | 4.2 | 4.1 | 9.0 | 4.2 | 14.4 | 21.6 |

| WE | | | | | | | | | EE | | | | | |
|---|---|---|---|---|---|---|---|---|---|---|---|---|---|---|
| it | kea | lb | mt | nb | oc | pt | es | sv | cy | am | be | bg | cs | et |
| 4.5 | 7.3 | 6.7 | 7.3 | 10.2 | 20.9 | 3.9 | 3.2 | 7.3 | 14.6 | 18.9 | 7.4 | 6.5 | 4.3 | 7.5 |
| 4.5 | 7.3 | 6.7 | 7.2 | 10.1 | 21.9 | 3.9 | 3.3 | 7.2 | 14.5 | 19.0 | 7.4 | 7.9 | 4.3 | 7.6 |

| EE | | | | | | | | | | CMN | | | | |
|---|---|---|---|---|---|---|---|---|---|---|---|---|---|---|
| ka | lv | lt | mk | pl | ro | ru | sr | sk | sl | uk | ar | az | he | kk |
| 11.4 | 4.6 | 7.5 | 4.3 | 4.3 | 3.9 | 9.4 | 99.6 | 3.6 | 4.7 | 4.8 | 14.7 | 7.4 | 17.6 | 3.9 |
| 11.5 | 4.6 | 7.5 | 4.3 | 4.3 | 4.0 | 9.4 | 99.6 | 3.6 | 4.7 | 4.8 | 14.8 | 8.7 | 17.3 | 4.9 |

| CMN | | | | | | | | | | | SSA | | | |
|---|---|---|---|---|---|---|---|---|---|---|---|---|---|---|
| ky | mn | ps | fa | ckb | tg | tr | uz | af | am | ff | lg | ha | ig | kam |
| 7.4 | 12.2 | 15.7 | 6.2 | 89.8 | 5.8 | 9.7 | 7.9 | 23.9 | 18.9 | 18.5 | 11.5 | 8.8 | 16.6 | 16.3 |
| 7.4 | 12.2 | 15.8 | 6.2 | 89.6 | 5.8 | 9.7 | 7.9 | 22.0 | 19.0 | 18.6 | 11.1 | 8.7 | 16.6 | 17.6 |

| SSA | | | | | | | | | | | | SA | | |
|---|---|---|---|---|---|---|---|---|---|---|---|---|---|---|
| ln | luo | nso | ny | om | sn | so | sw | umb | wo | xh | yo | zu | as | bn |
| 5.9 | 8.2 | 11.9 | 7.7 | 10.9 | 4.6 | 11.7 | 6.4 | 16.9 | 20.6 | 9.2 | 35.8 | 7.8 | 16.8 | 12.6 |
| 6.0 | 8.2 | 12.0 | 7.9 | 11.3 | 4.6 | 11.8 | 6.4 | 17.0 | 20.7 | 9.0 | 35.8 | 7.9 | 16.8 | 12.6 |

| SA | | | | | | | | | | | | SEA | | |
|---|---|---|---|---|---|---|---|---|---|---|---|---|---|---|
| gu | hi | kn | ml | mr | ne | or | pa | sd | ta | te | ur | my | ceb | tg |
| 11.2 | 15.0 | 7.4 | 12.1 | 8.5 | 9.8 | 13.1 | 99.9 | 93.1 | 11.6 | 11.0 | 9.1 | 21.3 | 6.3 | 5.8 |
| 11.3 | 15.0 | 7.4 | 12.1 | 8.5 | 9.9 | 13.0 | 99.9 | 93.2 | 11.6 | 11.0 | 9.2 | 21.4 | 6.3 | 5.8 |

| SEA | | | | | | | | CJK | | | ALL |
|---|---|---|---|---|---|---|---|---|---|---|---|
| id | jv | km | lo | ms | mi | th | vi | yue | cmn | ja | ko |
| 4.5 | 6.3 | 24.1 | 18.9 | 4.7 | 8.1 | 23.6 | 16.8 | 83.8 | 83.2 | 98.1 | 20.1 | **15.9** |
| 4.5 | 6.2 | 24.1 | 18.9 | 4.6 | 8.2 | 23.6 | 16.9 | 83.9 | 83.3 | 98.0 | 20.0 | **16.0** |

## 4. Conclusion

This paper introduces FLEURS-R, a speech restoration-applied version of the multilingual parallel corpus FLEURS. As this new corpus maintains N-way parallel property, it can be used for TTS as well as other speech generation tasks such as speech-to-speech translation, voice conversion, and speech retrieval. Through CERs computed by the Maestro-U ASR model and 5-scale naturalness MOS estimated by the SQuID model, we show that FLEURS-R data has better naturalness than the original speech while accurately maintaining its semantic content. Furthermore, the baseline TTS models built on this new dataset demonstrates that it was useful to build a multilingual TTS model. This improved corpus can enable significant progress towards building speech generation applications for everyone, including zero-shot and few-shot TTS in many languages.

# 5. References

[1] N. Chen, Y. Zhang, H. Zen, R. J. Weiss, M. Norouzi, and W. Chan, "WaveGrad: Estimating gradients for waveform generation," in *Proc. ICLR*, 2021.

[2] V. Popov, I. Vovk, V. Gogoryan, T. Sadekova, and M. Kudinov, "Grad-TTS: A diffusion probabilistic model for text-to-speech," in *Proc. ICML*, 2021.

[3] N. Zeghidour, A. Luebs, A. Omran, J. Skoglund, and M. Tagliasacchi, "SoundStream: An end-to-end neural audio codec," *IEEE/ACM Trans. ASLP*, vol. 30, pp. 495–507, 2021.

[4] A. Défossez, J. Copet, G. Synnaeve, and Y. Adi, "High fidelity neural audio compression," *Trans. MLR*, 2022.

[5] Z. Borsos, R. Marinier, D. Vincent, E. Kharitonov, O. Pietquin, M. Sharifi, D. Roblek, O. Teboul, D. Grangier, M. Tagliasacchi *et al.*, "AudioLM: a language modeling approach to audio generation," *IEEE/ACM Trans. ASLP*, 2023.

[6] C. Wang, S. Chen, Y. Wu, Z. Zhang, L. Zhou, S. Liu, Z. Chen, Y. Liu, H. Wang, J. Li *et al.*, "Neural codec language models are zero-shot text to speech synthesizers," *arXiv:2301.02111*, 2023.

[7] E. Kharitonov, D. Vincent, Z. Borsos, R. Marinier, S. Girgin, O. Pietquin, M. Sharifi, M. Tagliasacchi, and N. Zeghidour, "Speak, read and prompt: High-fidelity text-to-speech with minimal supervision," *Trans. ACL*, vol. 11, pp. 1703–1718, 2023.

[8] E. Casanova, J. Weber, C. D. Shulby, A. C. Junior, E. Gölge, and M. A. Ponti, "YourTTS: Towards zero-shot multi-speaker TTS and zero-shot voice conversion for everyone," in *Proc. ICML*, 2022, pp. 2709–2720.

[9] Z. Guo, Y. Leng, Y. Wu, S. Zhao, and X. Tan, "PromptTTS: Controllable text-to-speech with text descriptions," in *Proc. ICASSP*, 2023, pp. 1–5.

[10] D. Yang, S. Liu, R. Huang, C. Weng, and H. Meng, "InstructTTS: Modelling expressive TTS in discrete latent space with natural language style prompt," *arXiv:2301.13662*, 2023.

[11] P. K. Rubenstein, C. Asawaroengchai, D. D. Nguyen, A. Bapna, Z. Borsos, F. d. C. Quitry, P. Chen, D. E. Badawy, W. Han, E. Kharitonov *et al.*, "AudioPaLM: A large language model that can speak and listen," *arXiv:2306.12925*, 2023.

[12] A. Conneau, M. Ma, S. Khanuja, Y. Zhang, V. Axelrod, S. Dalmia, J. Riesa, C. Rivera, and A. Bapna, "FLEURS: Few-shot learning evaluation of universal representations of speech," in *Proc. SLT*. IEEE, 2023, pp. 798–805.

[13] Y. Koizumi, H. Zen, S. Karita, Y. Ding, K. Yatabe, N. Morioka, M. Bacchiani, Y. Zhang, W. Han, and A. Bapna, "LibriTTS-R: A restored multi-speaker text-to-speech corpus," *arXiv:2305.18802*, 2023.

[14] H. Zen, V. Dang, R. Clark, Y. Zhang, R. J. Weiss, Y. Jia, Z. Chen, and Y. Wu, "LibriTTS: A corpus derived from LibriSpeech for text-to-speech," in *Proc. Interspeech*, 2019, pp. 1526–1530.

[15] V. Pratap, Q. Xu, A. Sriram, G. Synnaeve, and R. Collobert, "MLS: A large-scale multilingual dataset for speech research," in *Proc. Interspeech*, 2020.

[16] "creative commons attribution 4.0 license (CC-BY 4.0)." [Online]. Available: https://creativecommons.org/licenses/by/4.0/

[17] F. S. Oliveira, E. Casanova, A. C. Júnior, A. S. Soares *et al.*, "CML-TTS: A multilingual dataset for speech synthesis in low-resource languages," *arXiv:2306.10097*, 2023.

[18] K. Prahallad, A. Vadapalli, N. Elluru, G. Mantena, B. Pulugundla, P. Bhaskararao, H. A. Murthy, S. King, V. Karaiskos, and A. W. Black, "The Blizzard Challenge 2013 – Indian language task," in *Blizzard Challenge workshop*, vol. 2013, 2013.

[19] Y. Koizumi, H. Zen, S. Karita, Y. Ding, K. Yatabe, N. Morioka, Y. Zhang, W. Han, A. Bapna, and M. Bacchiani, "Miipher: A robust speech restoration model integrating self-supervised speech and text representations," *arXiv:2303.01664*, 2023.

[20] Y.-A. Chung, Y. Zhang, W. Han, C.-C. Chiu, J. Qin, R. Pang, and Y. Wu, "w2v-BERT: Combining contrastive learning and masked language modeling for self-supervised speech pre-training," in *Proc. IEEE ASRU*, 2021.

[21] Y. Koizumi, S. Karita, S. Wisdom, H. Erdogan, J. R. Hershey, L. Jones, and M. Bacchiani, "DF-Conformer: Integrated architecture of Conv-TasNet and Conformer using linear complexity self-attention for speech enhancement," in *WASPAA*, 2021.

[22] Q. Wang, Y. Yu, J. Pelecanos, Y. Huang, and I. L. Moreno, "Attentive temporal pooling for Conformer-based streaming language identification in long-form speech," in *Odyssey*, 2022.

[23] Y. Jia, H. Zen, J. Shen, Y. Zhang, and Y. Wu, "PnG BERT: Augmented BERT on phonemes and graphemes for neural TTS," in *Proc. Interspeech*, 2021.

[24] Y. Koizumi, K. Yatabe, H. Zen, and M. Bacchiani, "WaveFit: An iterative and non-autoregressive neural vocoder based on fixed-point iteration," in *Proc. SLT*, 2023.

[25] Y. Zhang, W. Han, J. Qin, Y. Wang, A. Bapna, Z. Chen, N. Chen, B. Li, V. Axelrod, G. Wang, Z. Meng, K. Hu, A. Rosenberg, R. Prabhavalkar, D. S. Park, P. Haghani, J. Riesa, G. Perng, H. Soltau, T. Strohman, B. Ramabhadran, T. Sainath, P. Moreno, C.-C. Chiu, J. Schalkwyk, F. Beaufays, and Y. Wu, "Google USM: Scaling automatic speech recognition beyond 100 languages," *arXiv:2303.01037*, 2023.

[26] A. Pasad, J.-C. Chou, and K. Livescu, "Layer-wise analysis of a self-supervised speech representation model," in *ASRU*, 2021.

[27] Z. Chen, A. Bapna, A. Rosenberg, Y. Zhang, B. Ramabhadran, P. Moreno, and N. Chen, "Maestro-U: Leveraging joint speech-text representation learning for zero supervised speech asr," in *Proc. SLT*. IEEE, 2023, pp. 68–75.

[28] T. Sellam, A. Bapna, J. Camp *et al.*, "SQuId: Measuring speech naturalness in many languages," *arXiv:2210.06324*, 2022.

[29] T. Saeki, G. Wang, N. Morioka, I. Elias, K. Kastner, A. Rosenberg, B. Ramabhadran, H. Zen, F. Beaufays, and H. Shemtov, "Extending multilingual speech synthesis to 100+ languages without transcribed data," in *Proc. ICASSP*, 2024.

[30] A. Gulati, J. Qin, C.-C. Chiu, N. Parmar, Y. Zhang, J. Yu, W. Han, S. Wang, Z. Zhang, Y. Wu, and R. Pang, "Conformer: Convolution-augmented transformer for speech recognition," *Proc. Interspeech*, 2020.